\documentclass[10pt,twocolumn,letterpaper]{article}

\usepackage{cvpr}              
\usepackage{multirow}
\usepackage{amsmath}
\usepackage{makecell}
\usepackage{amsmath}
\usepackage{algorithmic}
\usepackage{algorithm}
\usepackage{amsmath}
\usepackage{listings}
\usepackage{colortbl}
\usepackage{booktabs} 
\usepackage{graphicx}
\usepackage{dsfont}
\usepackage{arydshln}
\usepackage{multirow}
\usepackage{multicol}
\usepackage{flushend}
\usepackage[misc]{ifsym}
\definecolor{Gray}{gray}{0.01}
\newcolumntype{a}{>{\columncolor{Gray}}c}

\usepackage[dvipsnames]{xcolor}

\definecolor{cvprblue}{rgb}{0.21,0.49,0.74}
\usepackage[pagebackref,breaklinks,colorlinks,citecolor=cvprblue]{hyperref}

\def\paperID{14661} 
\def\confName{CVPR}
\def\confYear{2024}

\title{Direct Distillation between Different Domains}

\author{
	Jialiang Tang$^{12,3}$, Shuo Chen$^{5,*}$, Gang Niu$^{5}$, Hongyuan Zhu$^{4}$, \\Joey Tianyi Zhou$^{4}$, Chen Gong$^{1,2,3,}$\thanks{Corresponding authors: Chen Gong (chen.gong@njust.edu.cn), Shuo Chen (shuo.chen.ya@riken.jp).}, Masashi Sugiyama$^{5,6}$ \vspace{0.3em} \\
	{\normalsize $^1$School of Computer Science and Engineering, Nanjing University of Science and Technology} \\
	{\normalsize $^2$Key Laboratory of Intelligent Perception and Systems for High-Dimensional Information of Ministry of Education} \\
{\normalsize $^3$Jiangsu Key Laboratory of Image and Video Understanding for Social Security} \\
{\normalsize $^4$Institute for Infocomm Research (I2R), the Agency for Science, Technology and Research (A$^{*}$STAR)} \\
{\normalsize $^5$Center for Advanced Intelligence Project, RIKEN} \\
{\normalsize $^6$The Graduate School of Frontier Sciences, The University of Tokyo}
}

\begin{document}
\maketitle
\begin{abstract}
Knowledge Distillation (KD) aims to learn a compact student network using knowledge from a large pre-trained teacher network, where both networks are trained on data from the same distribution. However, in practical applications, the student network may be required to perform in a new scenario (\emph{i.e.}, the target domain), which usually exhibits significant differences from the known scenario of the teacher network (\emph{i.e.}, the source domain). The traditional domain adaptation techniques can be integrated with KD in a two-stage process to bridge the domain gap, but the ultimate reliability of two-stage approaches tends to be limited due to the high computational consumption and the additional errors accumulated from both stages. To solve this problem, we propose a new one-stage method dubbed ``Direct Distillation between Different Domains" (4Ds). We first design a learnable adapter based on the Fourier transform to separate the domain-invariant knowledge from the domain-specific knowledge. Then, we build a fusion-activation mechanism to transfer the valuable domain-invariant knowledge to the student network, while simultaneously encouraging the adapter within the teacher network to learn the domain-specific knowledge of the target data. As a result, the teacher network can effectively transfer categorical knowledge that aligns with the target domain of the student network. Intensive experiments on various benchmark datasets demonstrate that our proposed 4Ds method successfully produces reliable student networks and outperforms state-of-the-art approaches.
\end{abstract}    
\begin{figure}[t]
	\centering
	\includegraphics[scale=0.269]{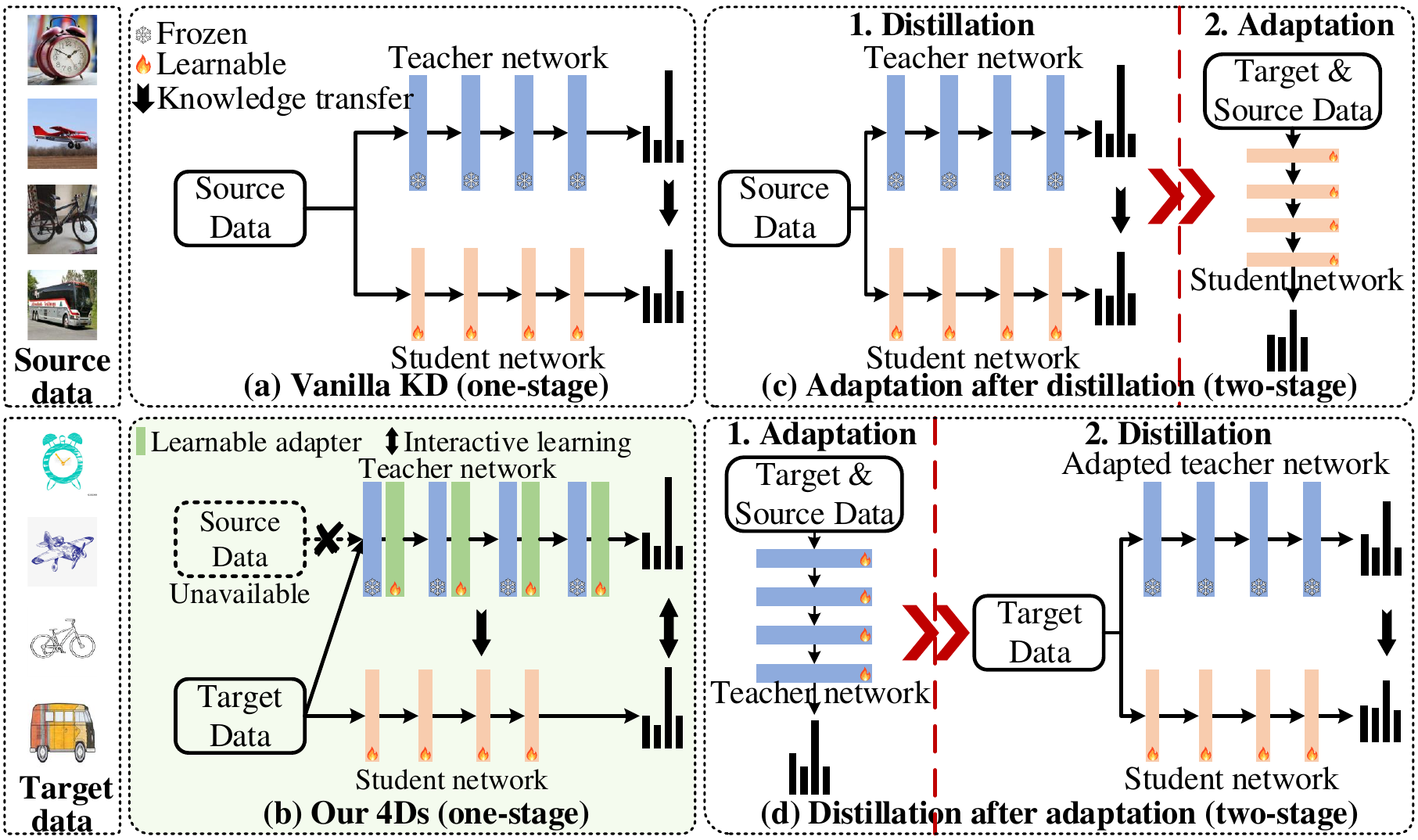}
	\caption{Comparison between existing Knowledge Distillation (KD) methods and our Direct Distillation between Different Domains (4Ds). (a) Vanilla KD transfers knowledge from a fixed pre-trained teacher network to the student network where the data is identically distributed. (b) Our 4Ds trains the student network on the target data using the teacher network (with the learnable adapters) trained on the source data. ``Adaptation after distillation'' in (c) first learns a student network on the source data via KD and then generalizes the student network on the target data via Domain Adaptation (DA). ``Distillation after adaptation'' in (d) first adapts the teacher network trained on the source data to the target data via DA and then utilizes the adapted teacher network to guide the student network training on the target data via KD.}
	\label{fig_varkd}
\end{figure}
\section{Introduction}
\label{sec:intro}
In recent years, Deep Neural Networks (DNNs) have achieved great success in many computer vision areas, such as image recognition~\cite{he2016deep,simonyan2014very} and object detection~\cite{girshick2014rich,girshick2015fast}. However, these powerful DNNs are usually infeasible for some resource-limited devices (\emph{e.g.}, smartphones or autonomous vehicles) due to their enormous computational and storage demands. To overcome this impediment, various model compression approaches have been proposed, mainly including network pruning~\cite{wang2021convolutional}, model quantization~\cite{choi2020data}, and Knowledge Distillation (KD)~\cite{hinton2015distilling}. Among these methods, KD has gained popularity for enhancing the performance of small student networks by transferring knowledge from large pre-trained teacher networks.

However, in practice, it is quite often that the acquired small student network should be capable of adapting to a new environment that differs from the one in which the original teacher network was trained. For example, in autonomous driving, one popular way is to train a well-performed yet large perception model in a simulation environment first, and eventually deploy it in real-world autonomous vehicles with limited computational and memory resources~\cite{huang2021model}. Nevertheless, due to the domain gap between the simulation environment (\emph{i.e.}, the source domain) and real-world driving scenarios (\emph{i.e.}, the target domain), existing KD methods cannot always perform well. These methods usually assume that the underlying data distributions for both the teacher network and student network are the same, and the direct knowledge transfer from the teacher network (well-trained in the source domain) to the student network (in the target domain) sometimes leads to poor performance on the downstream tasks. 

Intuitively, as shown in subfigures (c) and (d) of Fig.~\ref{fig_varkd}, there are two possible ways to mitigate such domain discrepancy in KD, namely: 1) First learning the student network in the source domain via KD, and then adapting it to the target domain (\emph{i.e.}, ``adaptation after distillation’’); and 2) First adapting the teacher network to the target domain, and then training the student network via KD (\emph{i.e.}, ``distillation after adaptation’’)~\cite{gao2022cross,islam2021dynamic}. However, both aforementioned approaches have clear shortcomings in terms of high computational complexity, potential error accumulation, and dependence on the source data. First, the two-stage approaches repeatedly iterate the teacher or student network, significantly increasing the computational costs. Second, neither distillation nor adaptation is fully reliable, so the two-stage framework may inevitably accumulate additional errors that further impair the model performance. Third, adapting the teacher or student network to an unfamiliar domain usually still requires the source data, which may be inaccessible due to privacy concerns. Therefore, if we want to effectively distill and adapt the knowledge of pre-trained models to our practical cases-of-interests with limited computational and storage resources, it is desirable to overcome the above disadvantages of two-stage methods. This motivates us to develop a one-stage end-to-end method that directly distills the critical knowledge from the source domain to the target domain without using the source data.

In this paper, we propose a novel method called ``\textbf{D}irect \textbf{D}istillation between \textbf{D}ifferent \textbf{D}omains’’ (4Ds), which straightforwardly trains a student network in the target domain by using only a well-trained teacher network in another different source domain. Specifically, as the teacher network contains both \textbf{\emph{domain-invariant}} knowledge (\emph{e.g.}, semantics) and \textbf{\emph{domain-specific}} knowledge (\emph{e.g.}, color and style), we design a novel Fourier transform based adapter and integrate such a learnable adapter into the teacher network to \textbf{\emph{decouple}} these two types of knowledge. After that, we retain the useful domain-invariant knowledge and transfer it to the student network via a fusion-activation mechanism. For the domain-specific information learned from the source domain, it is difficult to effectively transfer to the student network due to the discrepancies between the source and target domains. Therefore, we encourage the learnable adapter imposed on the teacher network to grasp new domain-specific information from the target domain, thereby improving the teacher's performance as a reliable supervisor for the student network. Consequently, the teacher network can capture reliable category relations compatible with the target domain, and successfully transfer them to the student network. The introduced adapter constitutes only 2\% of the teacher network parameters, making our method considerably more efficient than the two-stage approaches that repeatedly train the entire teacher or student network on source and target data. Furthermore, we directly extract the domain-invariant knowledge contained in the teacher network without relying on the source data, therefore effectively avoiding potential data privacy issues.

The contributions of this paper are summarized as:
\begin{enumerate}
	\itemsep=1pt
	\item We investigate a novel learning scenario of model compression between different domains, and it motivates us to propose a one-stage KD framework termed 4Ds, which directly trains a compact student network in a new target domain using a pre-trained teacher network from the source domain.
	\item We develop a new Fourier transform based adapter to decouple the domain-invariant knowledge and domain-specific knowledge contained in the teacher network, and then design a fusion-activation mechanism to transfer such valuable domain-invariant knowledge to the student network.
	\item Intensive experiments on multiple benchmarks demonstrate that our proposed 4Ds can outperform the two-stage methods and state-of-the-art knowledge transfer based approaches. 
\end{enumerate}
\section{Related Works}
\label{sec:related}
In this section, we review the relevant works, including knowledge distillation and domain adaptation. 
\subsection{Knowledge Distillation}
Conventional KD promotes a lightweight student network to acquire knowledge from a cumbersome teacher network to achieve satisfactory performance. In general, existing approaches assume that the data for the teacher network and student network follows the same distribution. According to the different knowledge being transmitted, existing KD approaches can be categorized into logit-based, feature-based, and relation-based methods. Logit-based methods~\cite{ba2014deep,hinton2015distilling} directly utilize the softmax-softened predictions to transfer the teacher-learned category similarities to the student network. In comparison, feature-based methods~\cite{chen2022knowledge,romero2014fitnets} derive the features from intermediate layers of the teacher network to train the student network, as these features of DNNs contain the richer information than their output predictions~\cite{zagoruyko2016paying}. More recently, relation-based methods~\cite{peng2019correlation, tung2019similarity} encourage the student network to learn intrinsic representations by utilizing the relations between examples in the teacher's representation space, which are calculated by various metric functions, such as the Euclidean distance and cosine similarity. Although these KD methods have achieved some promising results, they are usually ineffective in practice when the student network is trained in a new domain that has a different data distribution from the training domain of the teacher network. 

Departing from the conventional KD, some recently proposed data-free KD methods assume that the original training data of the teacher network is usually unavailable due to privacy concerns, so they train the student networks by model-generated data~\cite{chen2019data,do2022momentum} or web-collected real data~\cite{chen2021learning,tang2023distribution,fang2021mosaicking}. However, these methods still fail to train a reliable student network in a new domain that significantly differs from the training domain of the teacher network.
\subsection{Domain Adaptation}
Domain Adaptation (DA) leverages abundant labeled data or well-trained models from the source domain to enhance the model performance in the different target domains. In general, existing DA approaches bridge the distribution discrepancies between the source and target domains via instance reweighting~\cite{long2014transfer}, adversarial training~\cite{tzeng2017adversarial}, or knowledge transfer~\cite{islam2021dynamic}. Among them, knowledge transfer methods are rather simple and have been successfully applied to many fields, such as semantic segmentation~\cite{gao2022cross}, federated learning~\cite{gong2022preserving}, and few-shot learning~\cite{islam2021dynamic}. To improve the model generalize ability in the target domain, these methods transfer and align a variety of knowledge between models in the source domain and target domain, such as soft targets~\cite{li2023teacher}, gradient information~\cite{zhang2021matching}, sharpened predictions~\cite{islam2021dynamic}, and instance relations~\cite{gao2022cross,dong2021hrkd}. 

Note that although our approach shares some similar points with the DA methods that utilize knowledge transfer~\cite{gao2022cross,islam2021dynamic,zhang2021matching}, there are fundamental differences between our 4Ds and the existing methods: 1) (\emph{Research goal}) DA only adapts the model trained in the source domain to the target domain, whereas our approach further achieves model compression by training a small student network in the target domain; 2) (\emph{Employed strategies}) DA methods reduce the distribution discrepancies by aligning their distributions, while our approach improves the performance of the student network by extracting and transferring beneficial knowledge from the teacher network. To the best of our knowledge, we are the first to implement one-stage model compression under the learning scenario with different domains where the source-domain data is entirely unavailable. 
\begin{figure*}[t]
	\centering
	\includegraphics[scale=0.52]{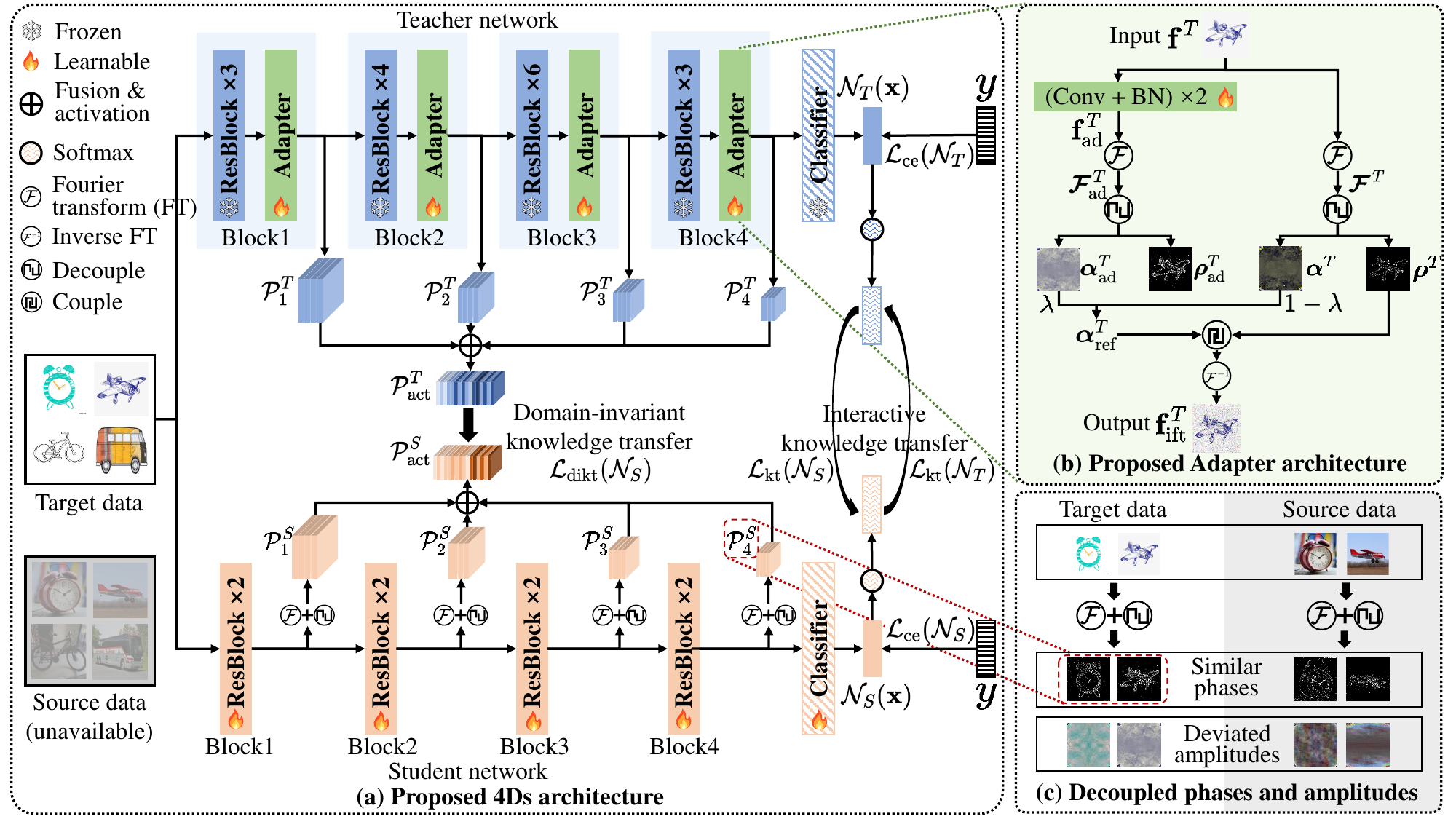}
	\caption{The diagram of our proposed 4Ds. (a) The teacher network $\mathcal{N}_{T}$ (ResNet34) and student network $\mathcal{N}_{S}$ (ResNet18) consist of four blocks, where each block further contains several ResBlocks. During training, both $\mathcal{N}_{T}$ and $\mathcal{N}_{S}$ interactively learn from the target data, where $\mathcal{N}_{T}$ is encouraged to produce accurate and useful category relations for $\mathcal{N}_{S}$ by updating its imposed adapters. Meanwhile, $\mathcal{N}_{S}$ is promoted to learn the valuable domain-invariant features as well as the reliable category relations from $\mathcal{N}_{T}$. (b) In our designed adapter, the input feature $\mathbf{f}^{T}$ is first fed into two learnable convolution layers to grasp the target-domain-specific knowledge. Subsequently, the original domain-specific knowledge is refurbished by mixing the amplitudes $\boldsymbol{\alpha}^{T}$ and $\boldsymbol{\alpha}^{T}_{\text{ad}}$, which are decoupled from the original $\mathbf{f}^{T}$ and adapted $\mathbf{f}^{T}_{\text{ad}}$, respectively. Finally, the output feature $\mathbf{f}^{T}_{\text{ift}}$ is recovered from the remained phase $\boldsymbol{\rho}^{T}$ from $\mathbf{f}^{T}$ and refurbished amplitude $\boldsymbol{\alpha}_{\text{ref}}^{T}$. (c) The input source images and target images are decoupled into phases and amplitudes by the Fourier transform and decoupling operations.}
	\label{fig_structure}
\end{figure*}
\section{Approach}
\label{sec:app}
In our problem setting, we aim to train a streamlined student network $\mathcal{N}_{S}$ in the new target domain using a pre-trained teacher network $\mathcal{N}_{T}$ from the source domain. Here, $\mathcal{N}_{T}$ is pre-trained on the source dataset $\mathcal{D}_{s}=\{(\mathbf{x}^{s}_{i},y^{s}_{i})\}_{i=1}^{|\mathcal{D}_{s}|}$ where $\mathbf{x}^{s}_{i}\in \mathcal{X}_{s}$, $y^{s}_{i}\in \mathcal{Y}_{s}$, and $\mathcal{N}_{S}$ is trained on the target dataset $\mathcal{D}_{t}=\{(\mathbf{x}^{t}_{i},y^{t}_{i})\}_{i=1}^{|\mathcal{D}_{t}|}$ where $\mathbf{x}^{t}_{i}\in \mathcal{X}_{t}$, $y^{t}_{i}\in \mathcal{Y}_{t}$, and ``$|\cdot|$'' symbolizes the data cardinality of the corresponding dataset throughout this paper. Although $\mathcal{D}_{s}$ and $\mathcal{D}_{t}$ share the same label space, their sample spaces differ from each other, \emph{i.e.}, the label spaces $\mathcal{Y}_{s}=\mathcal{Y}_{t}$ but the sample spaces $\mathcal{X}_{s}\neq \mathcal{X}_{t}$. Ideally, $\mathcal{N}_{S}$ should deliver satisfactory performance on $\mathcal{D}_{t}$ by mimicking the knowledge acquired by $\mathcal{N}_{T}$ that trained on $\mathcal{D}_{s}$. 

Specifically, $\mathcal{N}_{T}$ is initially pre-trained on $\mathcal{D}_{s}$, which embodies two distinct types of knowledge. The first one is \emph{\textbf{domain-specific}} knowledge, which refers to the inherent color and brightness of images in $\mathcal{D}_{s}$. These attributes significantly diverge from those present in $\mathcal{D}_{t}$ and are harmful to the training of $\mathcal{N}_{S}$ on $\mathcal{D}_{t}$. The second one is \emph{\textbf{domain-invariant}} knowledge, which pertains to the semantics of images across both $\mathcal{D}_{s}$ and $\mathcal{D}_{t}$. Such knowledge is shared between both two domains and is still beneficial for learning $\mathcal{N}_{S}$ on $\mathcal{D}_{t}$. Therefore, it is necessary to \emph{\textbf{decouple}} the domain-specific and domain-invariant knowledge embedded within the teacher network.

To overcome the challenges of KD under domain discrepancies with the above issues, we propose a one-stage end-to-end method called 4Ds. As shown in Fig.~\ref{fig_structure}, our 4Ds comprises three critical elements, including: 1) Knowledge adapter incorporating Fourier transform, which decomposes the spatial feature into domain-specific amplitude and domain-invariant phase via the Fourier transform; 2) Domain-invariant feature transfer, which distills the domain-invariant features from the teacher network to the student network via our designed fusion-activation mechanism; 3) Teacher-student interactive training, which alternatively trains the student network and learnable adapters in teacher network on the target data. Our 4Ds does not use any source data of the pre-trained teacher network. Therefore, for convenience in description, we symbolize the target data as $\mathcal{D}=\left\{\left(\mathbf{x}_i, y_i\right)\right\}_{i=1}^{|\mathcal{D}|}$ in the following discussions. Meanwhile, the notations with superscripts ``$T$'' and ``$S$'' denote that they are related to the teacher network $\mathcal{N}_{T}$ and student network $\mathcal{N}_{S}$, respectively.
\subsection{Knowledge Adapter Incorporating Fourier Transform}
\label{sec:aift}
In this subsection, we design an adapter incorporating the Fourier transform~\cite{cooley1969fast} to decouple the domain-specific and domain-invariant knowledge in the teacher network. Recent works~\cite{lee2023decompose,chen2021amplitude} leverage the Fourier transform~\cite{nussbaumer1982fast} to segregate spatial features from the intermediate layers of DNNs into amplitude and phase components and independently address these segregated components to enhance model ability. These studies suggest that the amplitude contains low-level information, such as color and brightness, which are prone to domain discrepancies. Meanwhile, the phase embodies high-level semantics that remain consistent across domains~\cite{xu2021fourier,lee2023decompose}. Based on this observation, we apply the Fourier transform incorporated in the adapter to segregate the amplitude and phase within the teacher network, rendering them the domain-specific feature and domain-invariant feature, respectively. 

As depicted in Fig.~\ref{fig_structure}, the teacher network comprises a total of $n$ blocks ($n$=4 for ResNet~\cite{he2016deep}), each following an efficient adapter consisting of only two 1$\times$1 convolutional layers. During training, all original components of the teacher network are frozen and only the integrated adapters are updated. This ensures the preservation of the domain-invariant knowledge in the original teacher network, which is beneficial for training the student network in the target domain. Conversely, the domain-specific knowledge that differs from the target domain is updated in the target domain by these learnable adapters.

For an input example $\mathbf{x}\in \mathcal{D}$ given to $\mathcal{N}_{T}$, the spatial feature $\mathbf{f}^{T} \in \mathbb{R}^{H \times W}$ ($H$ and $W$ denote the height and width of the feature map, respectively) produced by a specific block in $\mathcal{N}_{T}$ is first adapted by the learnable convolution layers in its adjacency adapter as follows:
\begin{equation}
\mathbf{f}^{T}_{\text{ad}} =  \text{BN}(\mathbf{W}_{\text{ad}}^{2}\delta(\text{BN}(\mathbf{W}_{\text{ad}}^{1}\mathbf{f}^{T}))).
\end{equation}
Here, $\mathbf{W}_{\text{ad}}^{1}\in \mathbb{R}^{\frac{C}{r_{\text{ad}}}\times C}$, and $\mathbf{W}_{\text{ad}}^{2}\in \mathbb{R}^{C\times  \frac{C}{r_{\text{ad}}}}$ are the weights for two convolutional layers; the channel dimension $C$ of input feature is omitted to simplify the description; $r_{\text{ad}}=4$ is a scaling parameter used to reduce computational costs; BN and $\delta$ denote the batch normalization layer and ReLU activation function, respectively. Then, the adapted feature $\mathbf{f}^{T}_{\text{ad}}\in \mathbb{R}^{H \times W}$ is converted to frequency feature $\boldsymbol{\mathcal{F}}_{\text{ad}}^{T}=\text{FT}(\mathbf{f}^{T}_{\text{ad}})$ by \textbf{F}ourier \textbf{t}ransform\footnote{Each Fourier coefficient $\boldsymbol{\mathcal{F}}_{\text{ad}}^{T}(u, v)$ is calculated as: 
	$\boldsymbol{\mathcal{F}}^{T}_{\text{ad}}(u, v)=\frac{1}{H W} \sum_{h=1}^{H} \sum_{w=1}^{W} \mathbf{f}_{\text{ad}}^{T}(h, w) e ^{-\mathrm{i} 2 \pi\left(\frac{u h}{H}+\frac{v w}{W}\right)}=\boldsymbol{\mathcal{F}}^{T}_{\text{ad\_real}}(u, v)+\mathrm{i} \boldsymbol{\mathcal{F}}^{T}_{\text{ad\_img}}(u, v),$
 where $\mathrm{i}$ is the imaginary unit, $\boldsymbol{\mathcal{F}}^{T}_{\text{ad\_real}}$ and $\boldsymbol{\mathcal{F}}^{T}_{\text{ad\_img}}$ are the real and imaginary parts of $\boldsymbol{\mathcal{F}}_{\text{ad}}^{T}$, respectively.}. Subsequently, the adapted amplitude $\boldsymbol{\alpha}^{T}_{\text{ad}}$ contains information specific to the target domain can be separated from $\boldsymbol{\mathcal{F}}_{\text{ad}}^{T}\in \mathbb{R}^{H \times W}$ leveraging the decoupling operation, namely:
\begin{equation}
(\boldsymbol{\alpha}^{T}_{\text{ad}}, \boldsymbol{\rho}^{T}_{\text{ad}}) = \text{Decouple}(\boldsymbol{\mathcal{F}}^{T}_{\text{ad}}),
\label{eq_dc1}
\end{equation}
where
\begin{equation}
\left\{
\begin{aligned}
\boldsymbol{\alpha}_{\text{ad}}^{T}(u, v)&=\sqrt{(\boldsymbol{\mathcal{F}}^{T}_{\text{ad\_real}}(u, v))^2+(\boldsymbol{\mathcal{F}}^{T}_{\text{ad\_img}}(u, v))^2}, \\
\boldsymbol{\rho}_{\text{ad}}^{T}(u, v)&=\arctan \frac{\boldsymbol{\mathcal{F}}^{T}_{\text{ad\_img}}(u, v)}{\boldsymbol{\mathcal{F}}^{T}_{\text{ad\_real}}(u, v)},
\end{aligned}
\right.
\end{equation}
respectively. Here, the amplitude $\boldsymbol{\alpha}_{\text{ad}}^{T}\in \mathbb{R}^{H \times W}$ and phase $\boldsymbol{\rho}_{\text{ad}}^{T}\in \mathbb{R}^{H \times W}$. Meanwhile, the original feature $\mathbf{f}^{T}$ is also decoupled to $\boldsymbol{\alpha}^{T}$ and $\boldsymbol{\rho}^{T}$ undergoes the same process.

Recent works~\cite{lee2023decompose,feng2023unsupervised} have suggested that completely eliminating the source-domain-specific information in DNNs may harm their performance in the target domain. In light of this, we refurbish the domain-specific feature in the teacher network by assembling both the $\boldsymbol{\alpha}^{T}_{\text{ad}}$ captured from the target domain and the original $\boldsymbol{\alpha}^{T}$ inherited by the teacher network using a learnable parameter $\lambda \in (0,1)$:
\begin{equation}
\boldsymbol{\alpha}_{\text{ref}}^{T} = \lambda\boldsymbol{\alpha}^{T}_{\text{ad}}+(1-\lambda)\boldsymbol{\alpha}^{T},
\end{equation}
where $\boldsymbol{\alpha}_{\text{ref}}^{T}\in \mathbb{R}^{H \times W}$ denotes the refurbished amplitude. Subsequently, the frequency feature $\boldsymbol{\mathcal{F}}_{\text{ref}}^{T}$ contains both valuable domain-invariant knowledge in the immutable $\boldsymbol{\rho}^{T}$ and domain-specific knowledge adaptable to the target domain in the refurbished $\boldsymbol{\alpha}_{\text{ref}}^{T}$ are composed as:
\begin{equation}
\boldsymbol{\mathcal{F}}_{\text{ref}}^{T}=\text{Couple}(\boldsymbol{\alpha}_{\text{ref}}^{T},\boldsymbol{\rho}^{T})=\boldsymbol{\alpha}_{\text{ref}}^{T} \cos (\boldsymbol{\rho}^{T})+\ \boldsymbol{\alpha}_{\text{ref}}^{T} \sin (\boldsymbol{\rho}^{T}) .
\end{equation}
Finally, the frequency feature $\boldsymbol{\mathcal{F}}_{\text{ref}}^{T}$ is recovered as spatial feature $\mathbf{f}^{T}_{\text{ift}}=\text{IFT}(\boldsymbol{\mathcal{F}}_{\text{ref}}^{T})$ via the \textbf{i}nverse \textbf{F}ourier \textbf{t}ransform\footnote{Each pixel $\mathbf{f}_{\text{ift}}^{T}(h, w)$ is computed as: \\$\mathbf{f}_{\text{ift}}^{T}(h, w)=\frac{1}{U V} \sum_{u=1}^{U} \sum_{v=1}^{V} \boldsymbol{\mathcal{F}}_{\text{ref}}^{T}(u, v) e ^{\mathrm{i} 2 \pi\left(\frac{u h}{U}+\frac{v w}{V}\right)}$.
}. Afterward, $\mathbf{f}^{T}_{\text{ift}}\in \mathbb{R}^{H \times W}$ is input into the subsequent blocks of the teacher network to produce accurate predictions.

During the 4Ds training process, the features produced by each teacher network's block are separated into domain-invariant and domain-specific features by their adjacent adapter. As a result, we can transfer valuable domain-invariant features to improve the performance of the student network (as detailed in Section~\ref{sec:diit}) and also promote a teacher network easily accessible by the student network to facilitate knowledge transfer (as explained in Section~\ref{sec:ito4}). The structure analysis and pseudo-code for the adapter are provided in the \textbf{Supplementary Materials}.
\subsection{Domain-Invariant Feature Transfer}
\label{sec:diit}
The domain-invariant features decoupled from the teacher network contain useful information similar to the target data, which benefits DNN training on the target domain~\cite{xu2021fourier,lee2023decompose}. Therefore, we develop the fusion-activation strategy to facilitate the transfer of domain-invariant features from the teacher network to the student network. Initially, the fusion operation integrates the features extracted from various blocks. Subsequently, the activation operation maps these fused features as attention weights to control the feature transfer.

In the fusion operation, an input example $\mathbf{x} \in \mathcal{D}$ is passed into the teacher network to extract the domain-invariant features $\{\boldsymbol{\mathcal{P}}^{T}_{i}\}_{i=1}^n$ from $n$ adapters, where $\boldsymbol{\mathcal{P}}^{T}_{i} \in \mathbb{R}^{C_{i} \times H_{i} \times W_{i}}$ and $\boldsymbol{\mathcal{P}}^{T}_{i}[j,:,:]=\boldsymbol{\rho}^{T}_{i,j} \in \mathbb{R}^{H_{i} \times W_{i}}$. In DNNs, the size of features generally decreases as the network deepens, which results in a size mismatch between the features from different adapters. To fuse these features, all features in $\{\boldsymbol{\mathcal{P}}^{T}_{i}\}_{i=1}^n$ are scaled to match those in $\boldsymbol{\mathcal{P}}^{T}_{n}$ with the smallest size using average pooling and then concatenated as:
\begin{equation}
\boldsymbol{\mathcal{P}}^{T}_{\text{fuse}}=\text{Concatenate}(\text{AvgPool}(\{\boldsymbol{\mathcal{P}}^{T}_{i}\}_{i=1}^n)),
\label{eq_fuse}
\end{equation}
where $\boldsymbol{\mathcal{P}}^{T}_{\text{fuse}} \in \mathbb{R}^{M \times H_{n} \times W_{n}}$ and $M=\sum_{i=1}^{n}C_{i}$.

After the feature fusion, the activation operation is utilized to capture the global dependencies within the fused features $\boldsymbol{\mathcal{P}}^{T}_{\text{fuse}}$ as the channel-wise attention weights. The features in $\boldsymbol{\mathcal{P}}^{T}_{\text{fuse}}$ are first average pooled to the channel-wise statistics with the size of 1$\times$1 as follows:
\begin{equation}
\text{z}_{m}=\frac{1}{H_{n} \times W_{n}} \sum_{h=1}^{H_{n}} \sum_{w=1}^{ W_{n}} \boldsymbol{\mathcal{P}}^{T}_{\text{fuse}}[m,h,w],
\end{equation}
where $\boldsymbol{\mathcal{P}}^{T}_{\text{fuse}}[m,:,:]\in \mathbb{R}^{H_{n} \times W_{n}}$ represents the $m$-th feature in $\boldsymbol{\mathcal{P}}^{T}_{\text{fuse}}$. Then, $\mathbf{Z}=[\text{z}_{1},\cdots,\text{z}_{m},\cdots,\text{z}_{M}]^{\top}$ (``${\top}$'' denotes the transpose operation) is passed through two Fully-Connected (FC) layers to capture the relation between features in $\boldsymbol{\mathcal{P}}^{T}_{\text{fuse}}$ as the attention weights:
\begin{equation}
\mathbf{S}=\sigma(\mathbf{W}^{2}_{\text{act}} \delta(\mathbf{W}^{1}_{\text{act}} \mathbf{Z})),
\end{equation} 
where $\sigma(\cdot)$ denotes the sigmoid activation function, $\mathbf{W}^{1}_{\text{act}} \in \mathbb{R}^{\frac{M}{r_{\text{act}}} \times M}$, and $\mathbf{W}^{2}_{\text{act}} \in \mathbb{R}^{M \times \frac{M}{r_{\text{act}}}}$ are the weights of the two FC layers, and $r_{\text{act}}=4$. Based on the attention weights in $\mathbf{S}=[s_{1},\cdots,s_{m},\cdots,s_{M}]^{\top}$, $\forall s_{m} \in [0,1]$, the features in $\boldsymbol{\mathcal{P}}^{T}_{\text{fuse}}$ can be activated as:
\begin{equation}
\boldsymbol{\mathcal{P}}^{T}_{\text{act}}= \boldsymbol{\mathcal{P}}^{T}_{\text{fuse}}\odot \text{Expand}(\mathbf{S},(M, H_{n}, W_{n})),
\end{equation}
where the function $\text{Expand}(\mathbf{S}, (M, H_{n}, W_{n}))$ extends the dimension of $\mathbf{S}$ to $\mathbb{R}^{M \times H_{n} \times W_{n}}$. Concurrently, the features from each block of the student network without the adapter are directly converted to frequency features by the Fourier transform and decoupled into domain-invariant features as $\{\boldsymbol{\mathcal{P}}^{S}_{i}\}_{i=1}^n$. Then, $\{\boldsymbol{\mathcal{P}}^{S}_{i}\}_{i=1}^n$ is fused as $\boldsymbol{\mathcal{P}}^{S}_{\text{fuse}}$ via Eq.~\eqref{eq_fuse} and subsequently activated by the attention weights in $\mathbf{S}$:
\begin{equation}
\boldsymbol{\mathcal{P}}^{S}_{\text{act}}= \boldsymbol{\mathcal{P}}^{S}_{\text{fuse}}\odot \text{Expand}(\mathbf{S}, (M, H_{n}, W_{n})).
\end{equation}

Based on $\{\boldsymbol{\mathcal{P}}^{S}_{\text{act},i}\}_{i=1}^{|\mathcal{D}|}$ of $\mathcal{N}_{S}$ and $\{\boldsymbol{\mathcal{P}}^{T}_{\text{act},i}\}_{i=1}^{|\mathcal{D}|}$ of $\mathcal{N}_{T}$, we can distill the domain-invariant knowledge of the teacher network to the student network using the following \textbf{d}omain-\textbf{i}nvariant \textbf{k}nowledge \textbf{t}ransfer loss function $\mathcal{L}_{\text{dikt}}\left(\mathcal{N}_S\right)$:
\begin{equation}
\mathcal{L}_{\text{dikt}}\left(\mathcal{N}_S\right)=\frac{1}{|\mathcal{D}|} \sum_{i=1}^{|\mathcal{D}|} \mathcal{H}_{\text {mse}}\left(\boldsymbol{\mathcal{P}}^{S}_{\text{act},i}, \boldsymbol{\mathcal{P}}^{T}_{\text{act},i}\right),
\label{eq_dikt}
\end{equation}
where $\mathcal{H}_{\text {mse}}$ represents the \textbf{m}ean \textbf{s}quare \textbf{e}rror that estimates the differences between the two input features. During distillation, the student network is promoted to learn expressive channels among domain-invariant features contained by the teacher network under the guidance of the attention weights, thereby enhancing its performance in the target domain. In some cases, this feature transfer process might be hindered by the feature dimension mismatched between the $\mathcal{N}_T$ and $\mathcal{N}_S$. In the \textbf{Supplementary Materials}, we address this problem through a feature mapping operation and provide a pseudo-code of the fusion-activation mechanism.
\begin{algorithm}[t]
	\caption{Direct Distillation between Different Domains}
	\begin{algorithmic}[1]
		\REQUIRE Pre-trained teacher network $\mathcal{N}_{T}$ in the source domain, target data $\mathcal{D}=\left\{\left(\mathbf{x}_{i}, y_{i}\right)\right\}_{i=1}^{|\mathcal{D}|}$, trade-off parameters $\beta$ and $\gamma$.
		\STATE Initialize and integrate the learnable adapters into $\mathcal{N}_{T}$;
		\STATE Initialize the small student network $\mathcal{N}_{S}$;
		\REPEAT 
		\STATE Employ $\mathcal{N}_{T}$, $\mathcal{N}_{S}$ to obtain their predictions $\{\mathcal{N}_{T}{(\mathbf{x}_{i})}\}_{i=1}^{|\mathcal{D}|}$, $\{\mathcal{N}_{S}{(\mathbf{x}_{i})}\}_{i=1}^{|\mathcal{D}|}$, and domain-invariant features $\{\boldsymbol{\mathcal{P}}_{\text{act}, i}^T\}_{i=1}^{|\mathcal{D}|}$, $\{\boldsymbol{\mathcal{P}}_{\text{act}, i}^S\}_{i=1}^{|\mathcal{D}|}$;
		\STATE Calculate cross-entropy losses $\mathcal{L_{\mathrm{ce}}}(\mathcal{N}_{T})$ and $\mathcal{L_{\mathrm{ce}}}(\mathcal{N}_{S})$;
		\STATE Calculate knowledge transfer losses $\mathcal{L_{\mathrm{kt}}}(\mathcal{N}_{T})$ and $\mathcal{L_{\mathrm{kt}}}(\mathcal{N}_{S})$ via Eq.~\eqref{eq_ktt};
		\STATE Calculate domain-invariant knowledge transfer loss $\mathcal{L_{\mathrm{dikt}}}(\mathcal{N}_{S})$ via Eq.~\eqref{eq_dikt};
		\STATE Update the learnable adapters in $\mathcal{N}_{T}$ via Adam;
		\STATE Update the student network $\mathcal{N}_{S}$ via SGD;
		\UNTIL {convergence}
		\ENSURE Lightweight student network in the target domain.
	\end{algorithmic}
	\label{alg_1}
\end{algorithm}
\begin{table*}[t]
	\resizebox{\linewidth}{!}{
		\begin{tabular}{c|cc|cccccccccccc|ccc}
			\hline
			Dataset                         & Teacher    & Student   & A2C    & A2P    & A2R    & C2A    & C2P    & C2R    & P2A    & P2C    & P2R    & R2A    & R2C    & R2P  &\textbf{Avg. (4Ds)} & Avg.$^{T}$ & Avg.$^{S}$   \\ \hline
			\multirow{4}{*}{Office-Home~\cite{venkateswara2017deep}}    & ResNet34   & ResNer18  & 81.41 & 92.12 & 84.86 & 71.81 & 92.11 & 85.20 & 72.42 & 81.33 & 85.44 & 72.63 & 81.67 & 92.22 & \textbf{82.77} (\textcolor{green}{5.03}) & 82.63 & 77.74\\ 
			& ResNeXt101 & ResNeXt50 & 84.10 & 95.49 & 89.44 & 83.74 & 95.60 & 89.22 & 84.74 & 84.87 & 89.22 & 84.36 & 84.76 & 95.49 & \textbf{88.42} (\textcolor{green}{3.71})& 87.99 & 84.71\\ 
			& VGGNet16   & VGGNet13  & 80.52 & 90.76 & 80.61 & 64.99 & 89.63 & 81.88 & 65.63 & 80.42 & 81.42 & 65.02 & 79.83 & 90.76 & \textbf{79.29} (\textcolor{green}{2.44})& 80.51 & 76.85\\ 
			& ResNet34   & VGGNet13  & 81.56 & 93.13 & 84.63 & 68.93 & 93.13 & 84.51 & 69.13 & 81.55 & 84.51 & 69.96 & 81.55 & 93.47 & \textbf{82.17} (\textcolor{green}{5.32})& 82.63 & 76.85\\ \hline \hline
			Dataset                         & Teacher    & Student   & A2C    & A2D    & A2W    & C2A    & C2D    & C2W    & D2A    & D2C    & D2W    & W2A    & W2C    & W2D & \textbf{Avg. (4Ds)} & Avg.$^{T}$ & Avg.$^{S}$ 
			\\ \hline
			\multirow{4}{*}{Office-Caltech~\cite{gong2012geodesic}} & ResNet34   & ResNer18  & 96.44 &100.00 & 100.00 & 97.39 & 100.00 & 98.31 & 97.40 & 96.88 & 100.00 & 97.15 & 97.38 & 100.00 & \textbf{98.33} (\textcolor{green}{2.14})& 98.64 & 96.19\\ 
			& ResNeXt101 & ResNeXt50 & 99.55 & 100.00 & 100.00 & 97.39 & 100.00 & 100.00 & 97.91 & 99.56 & 100.00 & 97.35 & 99.55 & 100.00 & \textbf{99.27} (\textcolor{green}{1.56})& 99.35 & 97.71\\ 
			& VGGNet16   & VGGNet13  &95.55	&96.87	&98.30	&97.91	&100.00	&98.30	&97.39	&96.44	&100.00	&97.35	&95.55	&100.00 & \textbf{97.80} (\textcolor{green}{1.64})& 98.16 & 96.16\\ 
			& ResNet34   & VGGNet13  & 96.44 & 100.00 & 100.00 & 96.91 & 100.00 & 98.30 & 97.91 & 96.00 & 100.00 & 97.11 & 96.44 & 100.00 & \textbf{98.25} (\textcolor{green}{2.09})& 98.64 & 96.16\\ \hline \hline
			Dataset                         & Teacher    & Student   & A2C    & A2P    & A2S    & C2A    & C2P    & C2S    & P2A    & P2C    & P2S    & S2A    & S2C    & S2P  & \textbf{Avg. (4Ds)} & Avg.$^{T}$ & Avg.$^{S}$   \\ \hline
			\multirow{4}{*}{PACS~\cite{zhou2020deep}}           & ResNet34   & ResNer18  & 98.09 & 99.40 & 97.07 & 97.12 & 99.10 & 96.56 & 96.87 & 98.30 & 96.69 & 95.87 & 97.73 & 99.10 & \textbf{97.65} (\textcolor{green}{2.98})& 96.55 & 94.67\\  
			& ResNeXt101 & ResNeXt50 & 98.57 & 99.70 & 96.94 & 98.29 & 99.70 & 97.45 & 97.56 & 98.93 & 97.33 & 97.32 & 99.14 & 99.70 & \textbf{98.38} (\textcolor{green}{3.31})& 97.64 & 95.07\\  
			& VGGNet16   & VGGNet13  & 96.80 & 98.80 & 96.05 & 95.12 & 98.80 & 96.69 & 95.85 & 97.23 & 95.80 & 94.88 & 97.01 & 98.80 & \textbf{96.81} (\textcolor{green}{3.31})& 95.29 & 93.50\\ 
			& ResNet34   & VGGNet13  & 94.45 & 99.40 & 96.43 & 93.90 & 99.40 & 96.43 & 95.41 & 97.45 & 96.56 & 94.90 & 96.45 & 99.40 & \textbf{96.68} (\textcolor{green}{3.18})& 96.55 & 93.50\\ \hline \hline
			Dataset                         & Teacher    & Student   & C2P    & C2R    & C2S    & P2C    & P2R    & P2S    & R2C    & R2P    & R2S    & S2C    & S2P    & S2R & \textbf{Avg. (4Ds)} & Avg.$^{T}$ & Avg.$^{S}$    \\ \hline
			\multirow{4}{*}{DomainNet~\cite{peng2019moment}}      & ResNet34   & ResNer18  & 67.61 & 80.63 & 66.35 & 74.49 & 80.50 & 66.05 & 74.77 & 68.53 & 66.25 & 75.18 & 68.73 & 80.46 & \textbf{72.46} (\textcolor{green}{2.64})& 72.82 & 69.82\\  
			& ResNeXt101 & ResNeXt50 & 73.99 & 84.89 & 71.23 & 78.63 & 84.77 & 71.40 & 78.84 & 74.22 & 71.89 & 78.75 & 73.89 & 84.84 & \textbf{77.27} (\textcolor{green}{3.62})& 77.11 & 73.65\\ 
			& VGGNet16   & VGGNet13  & 67.82 & 79.79 & 64.68 & 73.52 & 79.98 & 64.76 & 73.91 & 67.67 & 64.85 & 74.11 & 67.71 & 79.81 & \textbf{71.55} (\textcolor{green}{3.40})& 70.54 & 68.15 \\
			& ResNet34   & VGGNet13  &67.54	&80.27	&66.18	&73.91	&80.3	&66.01	&74.3	&67.81	&66.04	&74.54	&67.52	&80.02 &\textbf{72.03} (\textcolor{green}{3.88})& 72.82 & 68.15 \\  \hline
	\end{tabular}}
	\caption{Classification accuracies (in \%) of the experiments on four DA datasets. The column ``Avg. (4Ds)'' provides the mean accuracy achieved by student networks trained via our 4Ds. The column ``Avg.$^{T}$''/``Avg.$^{S}$" shows the average accuracy of four subsets among the corresponding dataset produced by the teacher/student networks trained by standard Back-Propagation (BP). The performance improvement of our 4Ds compared with ``Avg.$^{S}$" is indicated in green font in the brackets. For the DA task ``A2C'', the teacher network is pre-trained in the Art domain and the student network is trained in the Clipart domain, and the same applies to other items.}
	\label{table_benchmark}
	\vspace{-5pt}
\end{table*}
\subsection{Interactive Training of 4Ds}
\label{sec:ito4}
We propose an interactive training framework to concurrently train both the teacher network and student network on the target data. During training, the teacher network, which possesses useful domain-invariant knowledge, can quickly generalize well in the target domain. Moreover, we allow the teacher network to train alongside the student network, which can stimulate a student-friendly teacher network~\cite{park2021learning}. As a result, the teacher network can provide accurate and easily digestible knowledge to the student network, therefore improving the distillation performance. 

Specifically, the input target data $\mathcal{D}=\{(\mathbf{x}_{i},y_{i})\}^{|\mathcal{D}|}_{i=1}$ is first processed by both $\mathcal{N}_T$ and $\mathcal{N}_S$, producing the corresponding outputs $\{\mathcal{N}_T(\mathbf{x}_{i})\}^{|\mathcal{D}|}_{i=1}$ and $\{\mathcal{N}_S(\mathbf{x}_{i})\}^{|\mathcal{D}|}_{i=1}$. Then, a well-performed and student-friendly teacher network is achieved by minimizing the following integrated function:
\begin{equation}
\mathcal{L}_{\mathrm{total}}(\mathcal{N}_T)=\mathcal{L}_{\mathrm{ce}}(\mathcal{N}_T) + \beta\mathcal{L}_{\mathrm{kt}}(\mathcal{N}_T),
\label{eq_totalt}
\end{equation}
where $\mathcal{L_{\mathrm{ce}}}(\mathcal{N}_{T})=\frac{1}{|\mathcal{D}|} \sum_{i=1}^{|\mathcal{D}|}\mathcal{H}_{\mathrm{ce}}\left(\mathcal{N}_T\left(\mathbf{x}_i\right), y_i\right)$ is the \textbf{c}ross \textbf{e}ntropy loss function. $\mathcal{L}_{\mathrm{kt}}(\mathcal{N}_T)$ is the \textbf{k}nowledge \textbf{t}ransfer loss function and it is defined as follows:
\begin{equation}
\mathcal{L}_{\mathrm{kt}}\left(\mathcal{N}_T\right)=\frac{1}{|\mathcal{D}|} \sum_{i=1}^{|\mathcal{D}|}\mathcal{H}_{\mathrm{kl}}\left(\frac{\mathcal{N}_T\left(\mathbf{x}_i\right)}{\tau}, \frac{\mathcal{N}_S\left(\mathbf{x}_i\right)}{\tau}\right),
\label{eq_ktt}
\end{equation}
where $\tau=4$ is a default temperature parameter controlling the softening degree of soft targets $\frac{\mathcal{N}_T\left(\mathbf{x}_i\right)}{\tau}$ and $\frac{\mathcal{N}_S\left(\mathbf{x}_i\right)}{\tau}$, which contain the category relations between the interested categories in $\mathcal{D}$, and $\mathcal{H}_{\mathrm{kl}}$ represents the \textbf{K}ullback-\textbf{L}eibler divergence that measures the distribution discrepancies between two inputs. $\mathcal{L}_{\mathrm{ce}}(\mathcal{N}_T)$ and $\mathcal{L}_{\mathrm{kt}}(\mathcal{N}_T)$ motivate $\mathcal{N}_T$ to learn the knowledge specific to the target data and imitate the soft targets of $\mathcal{N}_S$, respectively, and their importances are tuned by a non-negative trade-off parameter $\beta$. 

Subsequently, the student network is trained on the target data while learning from the teacher network by the following total objective loss function:
\begin{equation}
\mathcal{L}_{\mathrm{total}}(\mathcal{N}_S)=\mathcal{L}_{\mathrm{ce}}(\mathcal{N}_S) + \beta\mathcal{L}_{\mathrm{kt}}(\mathcal{N}_S) + \gamma \mathcal{L}_{\text{dikt}}(\mathcal{N}_S),
\label{eq_totals}
\end{equation}
where $\gamma$ serves as another non-negative trade-off parameter, $\mathcal{L}_{\mathrm{kt}}(\mathcal{N}_S)$ encourages $\mathcal{N}_S$ to learn the readily learnable category relations from $\mathcal{N}_T$, and $\mathcal{L}_{\text{dikt}}(\mathcal{N}_S)$ (presented in Eq.~\eqref{eq_dikt}) promotes $\mathcal{N}_S$ to learn the valuable domain-invariant knowledge inherited by $\mathcal{N}_T$. Noticeably, $\mathcal{L}_{\text{dikt}}(\mathcal{N}_T)$ will compel $\mathcal{N}_T$ to mimic the domain-invariant features produced by $\mathcal{N}_S$, leading to dramatic disturbance of the domain-invariant features inherited by $\mathcal{N}_T$. Therefore, we reasonably exclude $\mathcal{L}_{\text{dikt}}(\mathcal{N}_T)$ from $\mathcal{L}_{\mathrm{total}}(\mathcal{N}_T)$ to preserve the valuable domain-invariant knowledge in $\mathcal{N}_T$.

As the training algorithm summarized in Alg.~\ref{alg_1}, during each training iteration, both the teacher network and student network are alternately updated by minimizing $\mathcal{L}_{\mathrm{total}}(\mathcal{N}_T)$ and $\mathcal{L}_{\mathrm{total}}(\mathcal{N}_S)$, respectively. The teacher network is promoted to capture domain-specific knowledge in the target domain and learn category relations that the student network can effortlessly learn. Subsequently, the well-preserved domain-invariant knowledge and appropriate category relations are transmitted to the student network to improve its performance in the target domain. As the training of the teacher network only requires a few calculations where the adapters encompass merely 2\% of the parameters of the entire teacher network, our proposed 4Ds can efficiently train a reliable student network in the target domain.
\section{Experiments}
\label{sec:exp}
In this section, we demonstrate the effectiveness of our proposed 4Ds on multiple popular benchmark datasets.\\
\textbf{Benchmark Datasets}: We perform intensive experiments on four widely used benchmark datasets, including: \textbf{Office-Caltech}~\cite{gong2012geodesic}, composed of 10 classes with approximately 2,500 images across four distinct domains (Amazon, Caltech, Dslr, Webcam); \textbf{Office-Home}~\cite{venkateswara2017deep}, contains 65 categories with 15,500 images belonging to four distinct domains (Art, Clipart, Product, Real); \textbf{PACS}~\cite{zhou2020deep}, consists of 7 classes with about 10,000 images from four domains (Art, Cartoon, Photo, Sketch); and \textbf{DomainNet}~\cite{peng2019moment}, currently the largest dataset for DA, which contains six domains with 345 categories, we follow~\cite{zhang2022divide,li2021cross} to select four domains (Clipart, Painting, Real, Sketch) for our experiments.\\
\textbf{Implementation Details}: All student networks in 4Ds employ Stochastic Gradient Descent (SGD) with weight decay of 5×10$^{-4}$ and momentum of 0.9 for optimization. Meanwhile, the adapters in the teacher network utilize Adam as the optimizer. All networks are trained over 120 epochs, the initial learning rates of the student network and adapters are 0.05 and 0.01, respectively, which are divided by 10 at the 40-th, 70-th, and 100-th epochs. Additionally, the trade-off parameters in Eq.~\eqref{eq_totalt} and Eq.~\eqref{eq_totals} are set to $\beta=1.0$ and $\gamma=0.1$, respectively. The parametric sensitivities of $\beta$ and $\gamma$ are investigated in the \textbf{Supplementary Materials}. 
\subsection{Experimental Results on Benchmark Datasets}
\label{sec:exp_res}
In this section, we conduct intensive experiments across four DA datasets to evaluate the effectiveness of our proposed 4Ds. Specifically, we consider four widespread teacher-student pairs, including ``ResNet34-ResNet18'', ``ResNeXt101-ResNeXt50'', ``VGGNet16-VGGNet13'', and ``ResNet34-VGGNet13''. The teacher networks are pre-trained on a specific subset (\emph{i.e.}, the source data) among a DA dataset and guide the student network training on the remaining subsets (\emph{i.e.}, the target data) of this DA dataset.

Table~\ref{table_benchmark} reports the corresponding classification results. We can observe that the student networks trained by our 4Ds consistently outperform ones trained on the target data by standard back-propagation. Moreover, the performance of compact student networks trained by our 4Ds is comparable to that of the complicated teacher network trained on the target data. These experimental results demonstrate that our 4Ds can effectively address the distribution discrepancies between the source domain and target domain without using any source data, leading to a superior student network. 
\subsection{Comparison with Domain Adaptation Methods}
\label{sec:exp_compare}
In this subsection, we compare our 4Ds with two-stage methods discussed in Section~\ref{sec:intro}, namely ``Distillation After Adaptation'' (``DAA'') and ``Adaptation After Distillation'' (``AAD''). During the adaptation stage, we employ Neighborhood Reciprocity Clustering (NRC)~\cite{yang2021exploiting} and Source HypOthesis Transfer (SHOT)~\cite{liang2021source} to generalize the teacher or student network from the source domain to the target domain. In the distillation stage, we follow VKD~\cite{hinton2015distilling} to transfer softened category probabilities produced by the teacher network to the student network. Moreover, we also compare our 4Ds with these DA methods using knowledge transfer, including Sharpened Predictions (SP)~\cite{islam2021dynamic}, Cross-domain Correlation Distillation (CCDistill)~\cite{gao2022cross}, Hierarchical Relational Knowledge Distillation (HRKD)~\cite{dong2021hrkd}, and Cross Domain Knowledge Distillation (CDKD)~\cite{li2023teacher}; and traditional KD methods, including VKD~\cite{hinton2015distilling}, Fitnets~\cite{romero2014fitnets}, Variational Information Distillation (VID)~\cite{ahn2019variational}, Similarity-Preserving Knowledge Distillation (SPKD)~\cite{tung2019similarity}, Semantic Calibration Knowledge Distillation (SemCKD)~\cite{chen2021cross}, and Softmax Regression Representation Learning (SRRL)~\cite{yang2020knowledge}. All codes and parameter settings are obtained from their official GitHub pages, and all methods use the teacher-student pair ResNet34-ResNet18 to train on four DA datasets.\\
\begin{table}[t]
	\resizebox{\linewidth}{!}{
		\begin{tabular}{cll|llll|l}
			\hline
			Type &Algorithm      &Required data          & Office-Home & Office-Caltech & PACS   & DomainNet & Avg.   \\ \hline
			\multicolumn{1}{c}{\multirow{2}{*}{\begin{tabular}[c]{@{}c@{}}Standard\\BP\end{tabular}}}	&Teacher         & Target                                   & 82.63          & 98.64       & 96.55  & 72.82     & 87.66 \\
			\multicolumn{1}{c}{} 	&Student         & Target                                   & 77.74          & 96.19       & 94.67  & 69.82     & 84.60 \\ \hline
			\multicolumn{1}{c}{\multirow{4}{*}{\begin{tabular}[c]{@{}c@{}}Two\\stage\end{tabular}}}	&DAA-NRC     & Target   & 78.31 (\textcolor{green}{0.57})         & 95.17 (\textcolor{red}{-1.02})      & 91.82 (\textcolor{red}{-2.85}) & 65.81 (\textcolor{red}{-4.01})    & 82.77 (\textcolor{red}{-1.83})\\
			\multicolumn{1}{c}{}	&DAA-SHOT    & Target  & 81.17 (\textcolor{green}{3.43})  & 95.75 (\textcolor{red}{-0.44})      & 95.67 (\textcolor{green}{1.00}) & 69.86 (\textcolor{green}{0.04})    & 85.61 (\textcolor{green}{1.01})\\
			\multicolumn{1}{c}{}	&AAD-NRC     & Target \& source                                   & 78.73 (\textcolor{green}{0.99})         & 96.10 (\textcolor{red}{-0.09})      & 91.89 (\textcolor{red}{-2.78}) & 66.68 (\textcolor{red}{-3.14})    & 83.35 (\textcolor{red}{-1.25})\\
			\multicolumn{1}{c}{}	&AAD-SHOT    & Target \& source                                    &79.57 (\textcolor{green}{1.83})          & 96.67 (\textcolor{green}{0.48})      & 95.61 (\textcolor{green}{0.94}) & 70.22 (\textcolor{green}{0.40})    & 85.51 (\textcolor{green}{0.91})\\ \hline
			\multicolumn{1}{c}{\multirow{4}{*}{\begin{tabular}[c]{@{}c@{}}Domain\\adaptation\end{tabular}}}	&SP~\cite{islam2021dynamic}    &Target \& source & 71.59 (\textcolor{red}{-6.15})        & 88.33 (\textcolor{red}{-7.86})     & 81.33 (\textcolor{red}{-13.34})& 55.81 (\textcolor{red}{-14.01})  & 74.26 (\textcolor{red}{-10.34})\\
			\multicolumn{1}{c}{}	&CCDistill~\cite{gao2022cross} &Target \& source & 79.38 (\textcolor{green}{1.64})        & 95.75 (\textcolor{red}{-0.44})     & 93.17 (\textcolor{red}{-1.50}) & 66.67 (\textcolor{red}{-3.15})    & 83.74 (\textcolor{red}{-0.86})\\
			\multicolumn{1}{c}{}	&HRKD~\cite{dong2021hrkd}      &Target \& source & 79.11 (\textcolor{green}{1.37})        & 96.12 (\textcolor{red}{-0.07})     & 95.18 (\textcolor{green}{0.51})& 68.37 (\textcolor{red}{-1.45})   & 84.69 (\textcolor{green}{0.09})\\
			\multicolumn{1}{c}{}	&CDKD~\cite{li2023teacher}    &Target & 79.29 (\textcolor{green}{1.55})        & 93.07 (\textcolor{red}{-3.12})     & 89.53 (\textcolor{red}{-5.14}) & 66.36 (\textcolor{red}{-3.46})   & 82.06 (\textcolor{red}{-2.54})\\ \hline
			\multicolumn{1}{c}{\multirow{7}{*}{\begin{tabular}[l]{@{}l@{}}Traditional\\knowledge\\distillation\end{tabular}}}	&VKD~\cite{hinton2015distilling}    &Target & 78.99 (\textcolor{green}{1.25})        & 94.37 (\textcolor{red}{-1.82})     & 90.02 (\textcolor{red}{-4.65})& 68.90 (\textcolor{red}{-0.92})   & 83.07 (\textcolor{red}{-1.53})\\
			\multicolumn{1}{c}{}	&Fitnets~\cite{romero2014fitnets} &Target & 78.52 (\textcolor{green}{0.78})        & 93.57 (\textcolor{red}{-2.62})     & 88.78 (\textcolor{red}{-5.89}) & 68.16 (\textcolor{red}{-1.66})   & 82.25 (\textcolor{red}{-2.35})\\
			\multicolumn{1}{c}{}	&VID~\cite{ahn2019variational}      &Target & 79.24 (\textcolor{green}{1.50})        & 94.37 (\textcolor{red}{-1.82})     & 90.05 (\textcolor{red}{-4.62})& 68.74 (\textcolor{red}{-1.08})   & 83.10 (\textcolor{red}{-1.50})\\
			\multicolumn{1}{c}{}	&SPKD~\cite{zhao2020multi}    &Target & 78.53 (\textcolor{green}{0.79})        & 93.72 (\textcolor{red}{-2.47})     & 87.61 (\textcolor{red}{-7.06})& 67.38 (\textcolor{red}{-2.44})   & 81.81 (\textcolor{red}{-2.79})\\ 
			\multicolumn{1}{c}{}	&SemCKD~\cite{chen2021cross}    &Target & 75.55 (\textcolor{red}{-2.19})        & 93.79 (\textcolor{red}{-2.40})     & 85.91 (\textcolor{red}{-8.76})& 66.70 (\textcolor{red}{-3.12})   & 80.48 (\textcolor{red}{-4.12})\\ 
			\multicolumn{1}{c}{}	&SRRL~\cite{yang2020knowledge}    &Target & 79.84 (\textcolor{green}{2.10})        & 93.76 (\textcolor{red}{-2.43})     & 90.02 (\textcolor{red}{-4.65})& 67.82 (\textcolor{red}{-2.00})   & 82.86 (\textcolor{red}{-1.74})\\  \hline
			Ours	& 4Ds                           &Target                      & \textbf{82.77} (\textbf{\textcolor{green}{5.03}})         & \textbf{98.33} (\textbf{\textcolor{green}{2.14}})      & \textbf{97.65} (\textbf{\textcolor{green}{2.98}})  & \textbf{72.46} (\textbf{\textcolor{green}{2.64}})     & \textbf{87.80} (\textbf{\textcolor{green}{3.20}})\\ \hline
	\end{tabular}}
	\caption{Average classification accuracies (in \%) of the student networks trained by various methods on 12 tasks for each dataset. Enhancements (or reductions) in performance attributed to applied methods are denoted in green (or red) font in brackets.}
	\label{table_compare}
	\vspace{-15pt}
\end{table}

Table~\ref{table_compare} presents the classification accuracies of the student networks trained by various compared methods and our 4Ds. Firstly, the student network trained by our one-stage 4Ds performs better than those trained by two-stage methods because our one-stage 4Ds can effectively circumvent the error accumulation problem inherent in two-stage methods. Meanwhile, our 4Ds also outperforms those DA methods utilizing knowledge transfer and traditional KD methods, which indicates that our 4Ds can effectively transfer the desired knowledge for the student network, and thus enhancing its performance.
\subsection{Ablation Studies}
\label{sec:exp_ab}
We select the teacher-student pair ResNet34-ResNet18 to estimate the key operations in 4Ds on the Office-Home dataset, and the results are shown in Table~\ref{table_ab}. The contributions of these key operations are analyzed as follows:\\
\textbf{1) Transferred knowledge} from the teacher network to the student network. We train a student network that only mimics the categorical knowledge or domain-invariant features from the teacher network. It can be observed that the performance of the student network without mimicking the categorical knowledge or domain-invariant features dramatically decreases, especially for domain-invariant features. Furthermore, the student network without learning both two types of knowledge performs the worst. These results clearly validate that both knowledge transferred in our 4Ds benefit the student network training.\\
\textbf{2) Weighting strategy} for domain-invariant features. We calculate ``Block-wise'', ``Channel-wise'', and ``Block-wise + Channel-wise'' weights (detailed in the \textbf{Supplementary Materials}) to transfer features from the teacher network to the student network. We can observe that 4Ds performs better than other strategies. This indicates that our fusion-activation mechanism, which captures the global dependencies of features from all network blocks, can effectively transfer the domain-invariant features. \\
\textbf{3) Training strategy} for our 4Ds. We do not use adapters, instead, the student networks are learned from a fixed teacher network and a learnable teacher network pre-trained in the source domain. The former cannot grasp the suitable domain-specific knowledge from the target domain, and the latter cannot retain the valuable domain-invariant knowledge learned from the source domain. We can observe that both of them show suboptimal performance compared with our 4Ds. These experimental results demonstrate that both preserving domain-invariant knowledge and updating domain-specific knowledge in our method are crucial for the student network to achieve satisfactory performance. Moreover, we also guide the training of the student network through a teacher network that uses adapters but without $\mathcal{L}_{\mathrm{ce}}(\mathcal{N}_T)$ or $\mathcal{L}_{\mathrm{kt}}(\mathcal{N}_T)$, and their performance is lower compared to the complete 4Ds. This indicates that a precise and student-friendly teacher network in the target domain is beneficial for training the student network.
\begin{table}[t]
	\centering
	\resizebox{\linewidth}{!}{
		\begin{tabular}{cl|cc}
			\hline
			\multicolumn{1}{c}{Type}                                                                             & Algorithm      & Avg. & Avg.$\downarrow$ \\ \hline
			
			\multicolumn{1}{c}{\multirow{3}{*}{\begin{tabular}[c]{@{}c@{}}Transferred knowledge\end{tabular}}} & w/o $\mathcal{L}_{\mathrm{kt}}(\mathcal{N}_S)$  & 81.80 & \textcolor{red}{0.97}      \\  
			\multicolumn{1}{c}{}                                                        & w/o $\mathcal{L}_{\mathrm{dikt}}(\mathcal{N}_S)$ & 80.25    & \textcolor{red}{2.52}      \\ 
			\multicolumn{1}{c}{}                                                                                 & w/o $\mathcal{L}_{\mathrm{kt}}(\mathcal{N}_S)$ + $\mathcal{L}_{\mathrm{dikt}}(\mathcal{N}_S)$     & 77.74    & \textcolor{red}{5.03}    \\
			\hline
			\multicolumn{1}{c}{\multirow{4}{*}{\begin{tabular}[c]{@{}c@{}}Weighting strategy\end{tabular}}}    & No weights & 80.07    & \textcolor{red}{2.70}     \\ 
			\multicolumn{1}{c}{}                                                                                 & Block-wise            & 81.39    & \textcolor{red}{1.38}      \\  
			\multicolumn{1}{c}{}                                                                                 & Channel-wise          & 81.44    & \textcolor{red}{1.33}      \\ 
			\multicolumn{1}{c}{}                                                                                 & Block-wise + Channel-wise  & 81.83    & \textcolor{red}{0.94}      \\ \hline
			\multicolumn{1}{c}{\multirow{4}{*}{\begin{tabular}[c]{@{}c@{}}Training strategy\end{tabular}}}     & Fixed teacher    		& 79.64    & \textcolor{red}{3.13}    \\ 
			\multicolumn{1}{c}{}                                                                                 & Learnable teacher    	& 78.55    & \textcolor{red}{4.22}     \\ 
			\multicolumn{1}{c}{}                                                                                 & w/o $\mathcal{L}_{\mathrm{ce}}(\mathcal{N}_T)$    	& 81.22    & \textcolor{red}{1.55}     \\ 
			\multicolumn{1}{c}{}                                                                                 & w/o $\mathcal{L}_{\mathrm{kt}}(\mathcal{N}_T)$    	& 81.86    & \textcolor{red}{0.91}    \\ \hline
			\multicolumn{1}{c}{4Ds}                                                                            & $\mathcal{L}_{\mathrm{total}}(\mathcal{N}_S)$                   & \textbf{82.77}    & \textcolor{red}{0.00}      \\ \hline
	\end{tabular}}
	\caption{Average classification accuracies (in \%) of the ablation experiments. The performance drop of each item compared with the complete 4Ds is indicated in red font in column ``Avg.$\downarrow$''.}
	\label{table_ab}
\vspace{-5pt}
\end{table}
\vspace{-5pt}
\section{Conclusion}
\label{sec:clu}
This paper proposes a new KD method called 4Ds to learn a compact student network in a new target domain from the teacher network trained within the familiar source domain. To our best knowledge, we are the first to perform one-stage model compression between different domains. Our proposed 4Ds adopts three key operations to address the domain discrepancy between source and target domains, namely: 1) Decoupling of domain-invariant and domain-specific knowledge in the teacher network; 2) Distilling the useful domain-invariant knowledge inherited by the teacher network to the student network; 3) Adapting domain-specific knowledge of the teacher network in the target domain to provide reliable knowledge to the student network. The main advantage of our method is that 4Ds is a one-stage end-to-end framework, so it successfully alleviates the error accumulation from the two stages of traditional approaches and avoids time-consuming iterations between these stages. Therefore, 4Ds can effectively solve the domain gap to train reliable student networks in the target domain by only using teacher networks in the source domain.

\section*{Acknowledgment}
C.G. was supported by NSF of China (No: 61973162), NSF of Jiangsu Province (No: BZ2021013), NSF for Distinguished Young Scholar of Jiangsu Province (No: BK20220080), the Fundamental Research Funds for the Central Universities (Nos: 30920032202, 30921013114), CAAI-Huawei MindSpore Open Fund, and “111” Program. M.S. was supported by the Institute for AI and Beyond, The University of Tokyo.
{
    \small
    \bibliographystyle{ieeenat_fullname}
    \bibliography{main}
}


\end{document}